\documentclass[11pt,a4paper,table,xcdraw]{article}
\usepackage[hyperref]{naaclhlt2019}
\usepackage{times}
\usepackage{latexsym}
\usepackage{amsmath}
\usepackage{tabularx}
\usepackage{booktabs}
\usepackage{url}
\usepackage{xcolor}
\usepackage{array}
\usepackage{graphicx}

\usepackage{multirow}
\usepackage[normalem]{ulem}
\useunder{\uline}{\ul}{}

\usepackage{verbatim}

\newcolumntype{M}[1]{>{\centering\arraybackslash}m{#1}}
\newcolumntype{Y}{>{\centering\arraybackslash}X}

\aclfinalcopy % Uncomment this line for the final submission
\title{The Steep Road to Happily Ever After: An Analysis of Current Visual Storytelling Models}

\author{Yatri Modi \and Natalie Parde \\ % Yatri
  Department of Computer Science \\
  University of Illinois at Chicago \\
  \{{\tt ymodi2, parde}\}{\tt @uic.edu}} 
\date{}

\begin{document}
\maketitle
\begin{abstract}
  Visual storytelling is an intriguing and complex task that only recently entered the research arena.  In this work, we survey relevant work to date, and conduct a thorough error analysis of three very recent approaches to visual storytelling.  We categorize and provide examples of common types of errors, and identify key shortcomings in current work.  Finally, we make recommendations for addressing these limitations in the future.
\end{abstract}

\section{Introduction}
\label{introduction}
\begin{figure}[t]
\includegraphics[width=7.5cm]{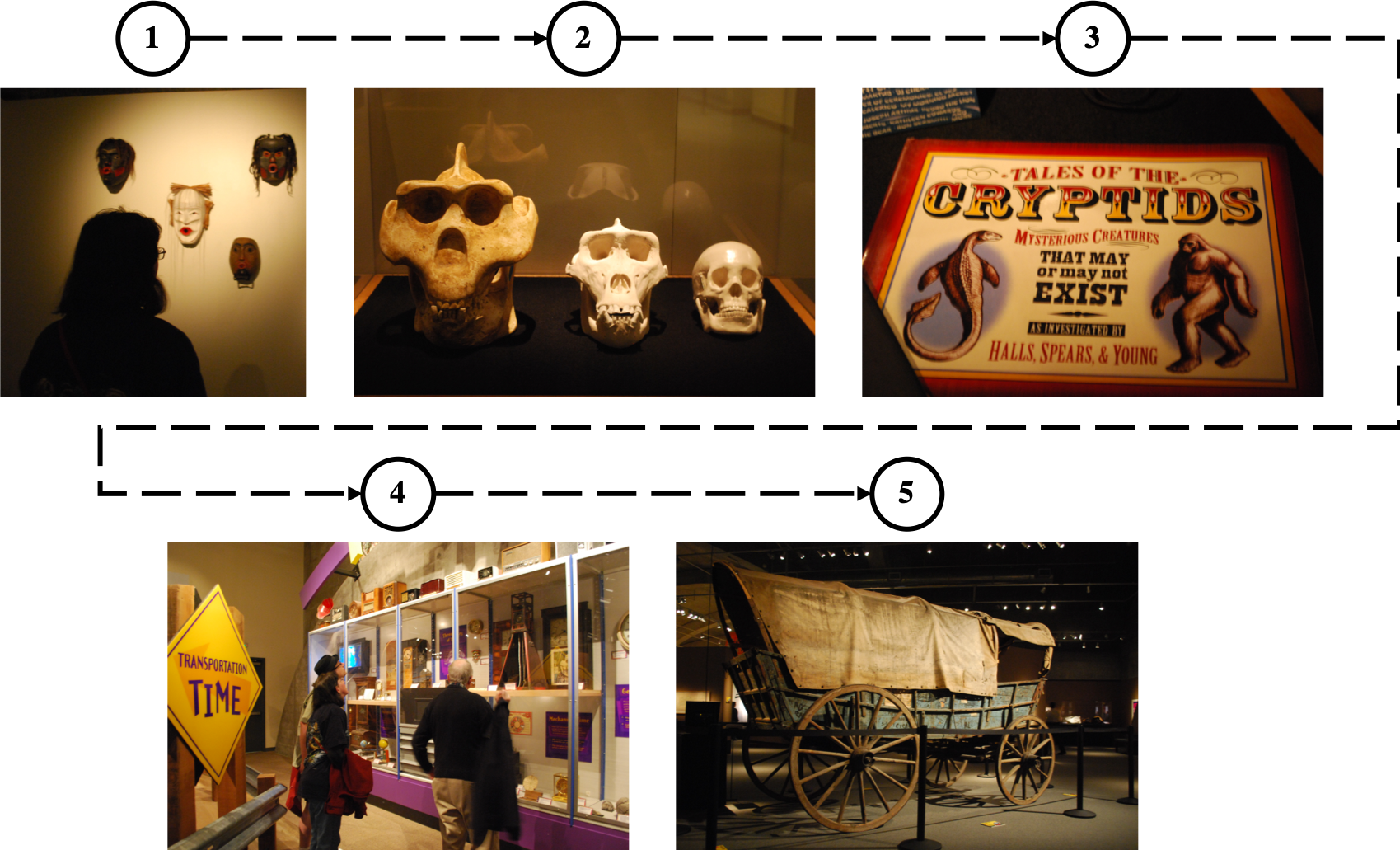}
\caption{A sequence of images from the VIST dataset.}
\label{image_sequence}
\end{figure}

Artificial intelligence continues to evolve, making it increasingly plausible to develop models that interpret vision and language in a humanlike manner.  A crucial element of such models is the capacity to not only match images with surface-level descriptions, but to infer deeper contextual meaning.  Recent literature has begun to refer to this task as \textit{visual storytelling}: the generation of a cohesive, sequential set of natural-language descriptions across multiple images \cite{vist}.  Visual storytelling is distinct from image captioning in that the text generated is oftentimes subjective, hinges on contextual image order, and typically employs more abstract and dynamic terms.  We illustrate the dichotomy between the two more concretely in terms of possible sets of sentences\footnote{Real samples (with punctuation and capitalization edited in some cases to increase readability) from the VIST dataset: \url{http://visionandlanguage.net/VIST/dataset.html}} for the images in Figure \ref{image_sequence}.

\textbf{Sentence Set 1:} (1) \textit{A woman looking at a collection of tribal masks on the wall.} (2) \textit{Three skulls of varying sizes ordered from largest to smallest.} (3) \textit{A top view of a book about mythical creatures.} (4) \textit{Three people standing in a store looking at the products.} (5) \textit{An old traveling wagon that is on display.}

\textbf{Sentence Set 2:} (1) \textit{I went to the natural history museum today.} (2) \textit{Their evolution display was very interesting.} (3) \textit{They had an area for cryptozoology.} (4) \textit{They also have a gift shop.} (5) \textit{My favorite was this real covered wagon from 200 years ago.}

The first is a set of traditional image captions, whereas the latter represents a visual story.  Note that the former presents factual descriptions of the images in isolation from one another.  The latter also describes the images, but places stronger emphasis on the development of a cohesive narrative underlying the image sequence.

High-performing visual storytelling approaches will enable growth for a variety of applications, many of which are associated with language understanding tasks.  They may also hold promise as a tool for assistive technology.  For instance, it is relatively common for users to upload large photo albums to social media platforms without including any image descriptions at all, making these albums inaccessible to those with sight impairments.  Visual storytelling could bridge this gap by automatically generating descriptive narratives for these albums.

Despite recent interest in visual storytelling, fueled by the 2018 Visual Storytelling Challenge,\footnote{\url{https://evalai.cloudcv.org/web/challenges/challenge-page/76/overview}} this research area is still quite nascent.  To date, no comprehensive review has been made of work on the task.  Such an analysis is necessary to spur additional research and recommend directions for future work.  Here, we fill this void, making the following contributions:
\begin{itemize}
    \item We catalogue existing models for visual storytelling, comparing and contrasting them with one another.
    \item We provide a performance comparison based on the original results (when publicly available) or re-implementations (when not).
    \item We categorize errors into distinct types and compile statistics indicating their frequencies within and across models.
    \item We make recommendations for addressing these errors in future visual storytelling models.
\end{itemize}

We discuss relevant prior work in Section \ref{related_work}, and describe the dataset used for visual storytelling tasks in Section \ref{data}.  In Section \ref{methods} we present an overview of the models included in our analysis, and in Section \ref{evaluation} we explain how these models were evaluated.  We conduct our comprehensive error analysis in Section \ref{error_analysis}, and make our recommendations based on the outcomes of this analysis in Section \ref{recommendations}.  We summarize these sections and report our final conclusions in Section \ref{conclusion}.

\section{Related Work}
\label{related_work}
We focus our analysis on methods employed by teams that participated in the 2018 Visual Storytelling Challenge.  The challenge required participants to make AI systems capable of generating human-like stories from a sequence of images as input. It had (1) an \textit{Internal Track} that constrained participants such that they could train only on data from the Visual Storytelling (VIST) Dataset, described further in Section \ref{data}, and use pretraining data only from any version of the ImageNet Large Scale Visual Recognition Challenge (ILSVRC)\footnote{A well-known annual competition that challenges researchers to solve a variety of large-scale object and image detection tasks \cite{ILSVRC15}:  \url{http://image-net.org/challenges/LSVRC/}.} and any version of the Penn Treebank;\footnote{A highly popular English-language part-of-speech tagset \cite{penn_treebank}: \url{https://catalog.ldc.upenn.edu/LDC99T42}.} and (2) an \textit{External Track} that allowed participants free reign when training, with the only requirement being that all training data be made publicly accessible if it was not already.  The challenge evaluated the quality of the generated stories using both an automatic metric (METEOR \cite{meteor}, described in further detail in Section \ref{metrics}) and human ratings corresponding to the following characteristics: (1) focus, (2) structure and coherence, (3) inclination to share, (4) likelihood of being written by a human, (5) visual grounding quality, and (6) level of detail.\footnote{Human judgements were solicited using Amazon Mechanical Turk (\url{https://www.mturk.com/}).} The winning team for the challenge was DG-DLMX \cite{DBLP:journals/corr/abs-1806-00738}.

We perform an in-depth error analysis of the work done by UCSB-NLP \cite{DBLP:journals/corr/abs-1804-09160}, SnuBiVtt \cite{DBLP:journals/corr/abs-1805-10973}, and DG-DLMX \cite{DBLP:journals/corr/abs-1806-00738} for the Visual Storytelling Challenge; these are the three teams who have released publicly available source code to date. We describe their models in further detail in Sections \ref{arel}-\ref{const}.  The other team participating in the challenge was NLPSA501 \cite{DBLP:journals/corr/abs-1805-11867}.  
%DG-DLMX \cite{DBLP:journals/corr/abs-1806-00738} developed a convolutional neural network (CNN) and long short-term memory (LSTM) encoder-decoder model that extended the image description approach originally proposed by \citeauthor{vinyals2015show} \shortcite{vinyals2015show}.  
NLPSA501 introduced a convolutional neural network (CNN) and gated recurrent unit (GRU) encoder-decoder model that incorporated an inter-sentence diverse beam search as a way to reduce redundancy in the generated stories.  We could not analyze their model's output as we did for those by UCSB-NLP, DG-DLMX and SnuBiVtt, due to the lack of available implementations or generated stories.

Outside of the Visual Storytelling Challenge, several other groups have explored the task of visual storytelling. \citeauthor{vist} \shortcite{vist} published the original paper introducing the visual storytelling task, comparing storytelling with image captioning. The authors used GRUs for both encoding the image and decoding the story. \citeauthor{W18-1503} \shortcite{W18-1503} defined a pipeline for visual storytelling consisting of Object Detection, Single-Image Inferencing, and Multi-Image Narration steps. \citeauthor{yu-bansal-berg:2017:EMNLP2017} \shortcite{yu-bansal-berg:2017:EMNLP2017} employed an alternate pipeline comprised of Album Encoder, Photo Selector, and Story Generator stages. \citeauthor{agrawal-EtAl:2016:EMNLP2016}'s \shortcite{agrawal-EtAl:2016:EMNLP2016} approach focuses on identifying proper sequences for existing story sentences, rather than on generating those sentences themselves.    Finally, \citeauthor{DBLP:journals/corr/JainAMSLS17} \shortcite{DBLP:journals/corr/JainAMSLS17} explored a phrase-based and syntax-based statistical machine translation approach as a vehicle for story generation using text but no images from the VIST dataset.  The approaches developed for the Visual Storytelling Challenge were designed to be improvements upon \citeauthor{vist}'s \shortcite{vist} model.  Although the approaches explored outside the challenge are not publicly available, we consider them when making our general recommendations.

The task of visual storytelling is still in its infancy, and to date there exists no comprehensive review of prior work in this area.  Our analysis fills this void, by summarizing relevant work in a shared context and providing concrete comparisons and example output when possible.  This allows us to identify core areas for improvement in future implementations, and recommend specific actions to address these current limitations.  Our hope is that this analysis can serve as a useful launchpad for other researchers aspiring to work in the visual storytelling domain.

\section{Data}
\label{data}
Most visual storytelling work to date has been trained and evaluated using the VIST Dataset \cite{vist}.  VIST is the first publicly available dataset for sequential vision-to-language tasks, and consists of sequences or ``albums'' of images wherein each image is paired with two types of captions; namely, descriptions of images in isolation (DII), and stories of images in sequence (SIS). The images were originally downloaded from Flickr (\url{https://www.flickr.com/}).  In total, the dataset comprises 10,117 Flickr albums containing 210,819 unique photos. 

Amazon Mechanical Turk (AMT) workers selected subsets of five images per album about which to write sequential, cohesive stories.  The dataset contains 50,200 story sequences overall; these are divided into subsets of 40,155 training, 4,990 validation and 5,055 testing stories.  Five written stories were collected per album.  Three standalone descriptions per image (DII, first defined above) were also collected separately using the image captioning interface used to build the COCO image caption dataset \cite{ms_coco}.  In both the stories and descriptions, all people names were replaced with generic MALE/FEMALE tokens, and all named entities were replaced with their entity type (e.g., location).  A small number of broken images were filtered from VIST by most research groups.  For concrete examples of DII and SIS from VIST, we refer readers to Figure \ref{image_sequence}, where Sentence Sets 1 and 2 (see Section \ref{introduction}) are from the DII and SIS subsets, respectively.

\section{Methods}
\label{methods}
We analyze three of the approaches submitted to the Visual Storytelling Challenge: AREL \cite{DBLP:journals/corr/abs-1804-09160}, GLACNet \cite{DBLP:journals/corr/abs-1805-10973} and Contextualize, Show and Tell \cite{DBLP:journals/corr/abs-1806-00738}.  We selected these approaches as the focus of our work for two reasons.  First, all were publicly available and well-documented, ensuring easy replicability.  Other existing visual storytelling models \cite{vist, agrawal-EtAl:2016:EMNLP2016, yu-bansal-berg:2017:EMNLP2017,  DBLP:journals/corr/abs-1805-11867, W18-1503} would have required reimplementation.  Doing so introduces the possibility of unintentionally crippling performance (e.g., when setting required but unreported parameters), which we wished to avoid.  Second, all were very recent models, representing the current state of the art in visual storytelling.  We summarize AREL, GLACNet, and Contextualize, Show and Tell in Sections \ref{arel}, \ref{glacnet}, and \ref{const}, and refer readers to the original papers for fuller detail.

\subsection{Adversarial Reward Learning (AREL)}
\label{arel}
AREL \cite{DBLP:journals/corr/abs-1804-09160} is an adversarial reinforcement learning approach that makes use of two models: a policy model, followed by a reward model.  The policy model is an encoder-decoder model utilizing a CNN-recurrent neural network (RNN) architecture, used to generate new stories. Specifically, a pre-trained CNN is fed a sequence of 5 images as input to extract high-level image features. These features are passed forward and further encoded as visual context vectors using bidirectional GRUs. The outputs of the encoder are then fed into a GRU-RNN decoder to generate sub-stories for the image sequence in parallel. The sub-stories are concatenated to form a single full story.  The CNN-based reward model is applied to every sub-story to compute its partial reward, and from the input sequence embeddings, n-gram features are extracted using convolution kernels of different sizes and passed through pooling layers. Image features are concatenated with these sentence representations and passed through a fully connected layer to obtain the final reward. To perform adversarial reward learning, the models were alternately optimized using stochastic gradient descent. The objective of the story generation policy was to maximize the similarity between a Reward Boltzmann distribution and itself. The first model optimized the policy to minimize the KL divergence \cite{kullback1951information} between itself and the Boltzmann Distribution, and the second model attempted to (a) minimize the KL divergence with the empirical distribution, and (b) maximize the KL divergence with the approximated policy distribution, with the objective of distinguishing between human and machine generated stories.

\citeauthor{DBLP:journals/corr/abs-1804-09160} \shortcite{DBLP:journals/corr/abs-1804-09160} demonstrated that AREL outperforms a generative adversarial network (GAN) model, a cross-entropy model, and other baselines and achieves state-of-the-art results across both automated and human metrics. The human metrics considered included both a Turing test (in which annotators attempted to guess which of two stories was written by a human) and pairwise comparisons measuring relevance, expressiveness, and concreteness.

\subsection{GLocal Attention Cascading Networks (GLACNet)}
\label{glacnet}
GLACNet \cite{DBLP:journals/corr/abs-1805-10973} also uses an encoder-decoder architecture, but it adds a hard attention mechanism which stresses feeding both the local image features and the overall context to the decoder as input. The image-specific features are extracted using a 152-layer residual network \cite{resnet}.  Those features are fed sequentially into a bidirectional LSTM, which then produces the global context vectors. The global context and local image features are combined to form \textit{glocal} vectors and passed through fully connected layers. The output is concatenated with word tokens and fed to the decoder (LSTM) as input. Thus, five glocal vectors for each image are fed into the decoder one after another, creating a cascading mechanism by passing the hidden state of one sentence generator as the initial hidden state of the next sentence generator.

To validate that all components of the GLACNet architecture contributed to the model's performance, \citeauthor{DBLP:journals/corr/abs-1805-10973} \shortcite{DBLP:journals/corr/abs-1805-10973} conducted an ablation study in which the cascading, global attention, local attention, and post-processing routines were removed one at a time, comparing perplexity and METEOR \cite{meteor} scores between conditions as well as with a standalone LSTM sequence-to-sequence (Seq2Seq) model and the full GLACNet model.  The full GLACNet model exhibited the best performance, and the other GLACNet-based models exhibited better performance than the LSTM Seq2Seq model, thereby verifying the utility of this approach.

\subsection{Contextualize, Show and Tell}
\label{const}
Contextualize, Show and Tell \cite{DBLP:journals/corr/abs-1806-00738} won the 2018 Visual Storytelling Challenge. The model uses an encoder LSTM to read in the image representations one by one for every image in a sequence. The image representations are generated using Inception V3 \cite{inception_v3}.
Five decoders, again LSTMs, then read in the image embedding as input. The first hidden state of each decoder is initialized using the last hidden state of the encoder to provide the model with global context.  \citeauthor{DBLP:journals/corr/abs-1806-00738} \shortcite{DBLP:journals/corr/abs-1806-00738} obtained the final story by concatenating the outputs of the model's five decoders.

As part of the Visual Storytelling Challenge, the model was evaluated on public and hidden test sets using both human evaluation and an automated metric (METEOR). METEOR scores of 30.88 and 31 were obtained on the public and hidden test sets, respectively.\footnote{\citeauthor{DBLP:journals/corr/abs-1806-00738} \shortcite{DBLP:journals/corr/abs-1806-00738} reported a METEOR score of 34.4 on the standard VIST test set.} Human evaluation scores were collected via Amazon Mechanical Turk. Crowd workers
evaluated six aspects of each story using a Likert scale. Each worker was asked to indicate the degree to which: 1) the story was focused, 2) the
story had good structure and coherence, 3) the worker would share the story, 4) the worker thought the story
was written by a human, 5) the story was visually
grounded, and 6) the story was detailed.  In summing the average scores received for each criterion, \citeauthor{DBLP:journals/corr/abs-1806-00738}'s \shortcite{DBLP:journals/corr/abs-1806-00738} model achieved a score of 18.498, whereas human-generated stories achieved a score of 23.596.

\section{Evaluation}
\label{evaluation}

\begin{table*}[t]
\centering
\small
\renewcommand{\arraystretch}{1.1}
\begin{tabular}{|p{1.8cm}|c|c|c|c|c|c|c|c|c|}
\hline
\textbf{Model}  & \textbf{METEOR} & \textbf{CIDEr} & \textbf{ROUGE-L}  &  \textbf{BLEU-1} & \textbf{BLEU-2} & \textbf{BLEU-3} & \textbf{BLEU-4} & \textbf{Perplexity}\\ \hline
\textit{AREL-s-50} & \textbf{34.9}  & \textbf{9.1}  & 29.4 & \textbf{62.9}  & \textbf{38.4} & \textbf{22.7} & \textbf{14.0}  & -\\ \hline
\textit{BLEU-RL}   & 34.6  & 8.9   & 29.0 & 62.1 & 38.0 & 22.6 & 13.9 & -      \\ \hline
\textit{CIDEr-RL}  & \textbf{34.9}  & 8.1 & \textbf{29.7}  & 61.9 & 37.8~ & 22.5~ & 13.8 & -      \\ \hline\hline

\textit{GLACNet} & 30.14                 & -                 & -                 & -    & - &-  & - &   \textbf{18.28}        \\ \hline\hline
\textit{Contextualize, Show and Tell} & 34.4                 & 5.1                 & 29.2                 & 60.1    & 36.5 &21.1  & 12.7 &  - \\ \hline 
\end{tabular}
\caption{Performance as reported in the source papers \cite{DBLP:journals/corr/abs-1804-09160,DBLP:journals/corr/abs-1805-10973}.  BLEU-RL, METEOR-RL, and CIDEr-RL were baseline reinforcement learning approaches using BLEU, METEOR, and CIDEr scores as their reward functions, respectively \cite{DBLP:journals/corr/abs-1804-09160}.}
\label{source_results}
\end{table*}

\begin{table*}
\centering
\small
\renewcommand{\arraystretch}{1.1}
\begin{tabular}{|p{1.8cm}|c|c|c|c|c|c|c|c|c|}
\hline
\textbf{Model}  & \textbf{METEOR} & \textbf{CIDEr} & \textbf{ROUGE-L}  &  \textbf{BLEU-1} & \textbf{BLEU-2} & \textbf{BLEU-3} & \textbf{BLEU-4} & \textbf{Perplexity}\\ \hline
\textit{AREL-s-50} & \textbf{35.2}  & \textbf{8.4}  & \textbf{29.9} & \textbf{61.9}  & \textbf{38.3} & \textbf{22.8} & \textbf{13.9}  & -\\ \hline

\textit{GLACNet} & 29.46                 & 3.7                 & 28.2                & 53.4    & 29.4 & 15.6  & 8.6 &   \textbf{19.51}        \\ \hline
\end{tabular}
\caption{Performance obtained when we ran AREL-s-50 and GLACNet, the two models for which we were able to obtain working implementations.}
\label{our_results}
\end{table*}

\subsection{Experimental Setup}
We trained and evaluated AREL according to the instructions provided in its publicly available Github repository.\footnote{\url{https://github.com/littlekobe/AREL}} However, we modified the source code slightly such that we were able to obtain the individual METEOR scores for each predicted story in the test set. This helped us in performing an in-depth error analysis of the generated stories and determining how well the automatic metrics were at scoring the stories. Training the model took around 2 weeks on a 3.5 GHz Intel Core i5 CPU with 16 GB RAM.\footnote{Extenuating circumstances limited our hardware resources in the midst of our AREL evaluation.  Training would have undoubtedly been quicker using GPUs, as was done in the original paper \cite{DBLP:journals/corr/abs-1804-09160}.} 

The GLACNet code is also publicly available.\footnote{ \url{https://github.com/tkim-snu/GLACNet}} We trained and evaluated the model using an NVIDIA Tesla P100 GPU instance on Google Cloud Platform. The model took one week to finish training. The original source code only provided an average METEOR score across all generated stories after testing.  Thus, we added code to produce the METEOR score for each story.  We will make all adapted source code publicly available online to ensure easy replicability.

The source code for Contextualize, Show and Tell is available online as well.\footnote{\url{https://github.com/dgonzalez-ri/neural-visual-storyteller}}  The authors personally sent us the generated stories, so we did not re-implement their model.  We have directly included their METEOR results in our evaluation.

\subsection{Evaluation Metrics}
\label{metrics}
Common metrics for evaluating visual storytelling models include METEOR \cite{meteor}, BLEU \cite{bleu}, CIDEr \cite{cider}, and ROUGE-L \cite{rouge-l}.  METEOR, the primary metric considered in the Visual Storytelling Challenge, calculates the alignment between the machine-generated hypotheses and the reference stories based on the exact, stem, synonym, and paraphrase matches between words and phrases.  While AREL was evaluated using METEOR as well as the other metrics, GLACNet was evaluated using only METEOR scores and measures of perplexity.  Contextualize, Show and Tell was also evaluated using only METEOR.  We generated scores for the remaining metrics as well for GLACNet and Contextualize, Show and Tell to aid our analysis.

\subsection{Results}
We observed slightly different results from those originally reported for the models included in our evaluation. We include both the originally-reported results and results based directly on original output files if available (Table \ref{source_results}) and our results from when we ran AREL and GLACNet (Table \ref{our_results}) in Tables \ref{source_results} and \ref{our_results}.  When we ran AREL and GLACNet, we collected scores for METEOR, CIDEr, ROUGE-L, BLEU-1, BLEU-2, BLEU-3, and BLEU-4, and found that  AREL outperformed GLACNet in all cases (Table 2).  We also found that based on \citeauthor{DBLP:journals/corr/abs-1804-09160}'s \shortcite{DBLP:journals/corr/abs-1804-09160} and \citeauthor{DBLP:journals/corr/abs-1806-00738}'s \shortcite{DBLP:journals/corr/abs-1806-00738} reported results and the additional metrics we computed for Contextualize, Show and Tell, the former outperformed the latter.

\section{Error Analysis}
\label{error_analysis}

We defined a threshold METEOR score of 25, with stories scoring below this threshold considered as serious errors.  This threshold was chosen following a manual assessment of the predicted stories, with METEOR \textless{} 25 representing a medium at which there existed both a sizable number of errors, and a sample of generated stories that were of noticeably low quality.  Stories having a METEOR score $\geq$50 were also analyzed for any anomalies (e.g., bad stories with high scores). 

Some metrics (CIDEr and BLEU-4) produced scores of 0 for many stories in both models. Upon manual analysis we found many of these stories to be sensible. Other work has confirmed that BLEU-3 and CIDEr scores do not correlate well with human evaluations \cite{DBLP:journals/corr/abs-1804-09160}.

We systematically analyzed each story in error and made notes indicating characteristics contributing to the error (including those that rendered the predicted stories to be completely meaningless or incoherent).  In the process, we also identified mechanisms by which those errors may be addressed in the future.  We compiled the errors into representative categories, which we define in Section \ref{error_categories} and exemplify in Table \ref{example_errors}.  We discuss these errors in fuller detail in Section \ref{error_discussion}.  In Section \ref{error_discussion} we also discuss some general errors from papers about other visual storytelling approaches for which we were unable to obtain full working implementations.

\subsection{Error Categories}
\label{error_categories}
\begin{table*}[t]
\centering
\small
\renewcommand{\arraystretch}{1.1}
\begin{tabular}{|p{2cm}|p{13cm}|}
\hline
\textbf{Error Type} & \textbf{Example} \\ \hline
\multirow{2}{*}{\begin{tabular}[c]{@{}l@{}}Grammatical\\ Errors\end{tabular}} & \textit{there was a lot of people at the convention center} . we saw a lot of interesting signs . there were a lot of people there . there were a lot of people there . we had a great time at the bar . (AREL) \\ \cline{2-2} 
 & the man was taking a walk on the sidewalk . he saw a lot of cool buildings . he saw a statue of a woman .\textit{ he was a big group of people} . he went to the museum . (GLACNet) \\ \hline
Contradictions & we went to the art gallery . we saw a lot of people there . \textit{the streets were empty . the streets were full of people }. this is a picture of a woman . (AREL) \\ \hline
\multirow{2}{*}{\begin{tabular}[c]{@{}l@{}}Repetitions \\within Story\end{tabular}}  &  \textit{ the bride and groom were very happy to be married }. the bride and groom were so happy to be married . \textit{the bride and groom were so happy to be married }. we all had a great time at the reception . they danced the night away . (AREL) \\ \cline{2-2}

& the family went to the zoo . they had a lot of fun . they were all very excited .\textit{ we had a great time . i had a great time .} (Contextualize, Show and Tell) \\ \hline

Repetitions within Sent. & it was a beautiful day for a trip to the beach . we took a trip to the beach . we went to the beach . the beach was beautiful . as \textit{ the sun went down , the sun went down }. (AREL) \\ \hline
Repetitive Subject & the water was calm and clear . the buildings were empty . the building was very tall . \textit{the architecture was amazing . the architecture was breathtaking }. (GLACNet) \\ \hline
Repetitive Sentence Structure & the city is very beautiful . the bridge is amazing . the water is so nice . the ferris wheel is very good . the view is spectacular . (GLACNet) \\ \hline
\multirow{2}{*}{\begin{tabular}[c]{@{}l@{}}Description \\in Isolation\end{tabular}}
&   \textit{this is a picture of a street }. it was a long drive . there was a lot of damage to the side of the road . \textit{this is a picture of a man }. after that we found a trail that was in the middle of the forest .  (BLEU-RL) \\ \cline{2-2}
& the flowers were very pretty the flowers were so beautiful . the flowers were beautiful . \textit{this is a picture of a column }. it was a very nice place to be .(Contextualize, Show and Tell)\\ \hline

Singular/Plural Disagreement & the resort was beautiful .\textit{ the beach was nice . the beaches were amazing }. the water was so calm . the food was delicious . (GLACNet) \\ \hline
Ghost Entities & the lady was smiling for the camera . she was excited to be there . she was having a good time . \textit{she was so happy to see \textbf{her}} . she was looking at the car  (GLACNet) \\ \hline
Personification & \textit{the plane was very excited to be at the location} . the first stop was the train station . the guide was also impressed with the organization organization . the students were able to see the exhibits from the city . the entire group was so happy to be there . (GLACNet) \\ \hline
\multirow{2}{*}{Absurdity} & \textit{the kitchen was a lot of work . here is a picture of a box . i had to take a picture of my work . we had to take a picture of the menu . i had a great time .} (AREL) \\ \cline{2-2} 
 & \textit{i bought a new car . this is a picture of a cat }. she was very excited . and i 'm so excited . this is my favorite gift . (GLACNet) \\ \hline
Incomplete Stories & i love to travel i had a great time . she is having a great time . we went to the city to see some of the people . i had a great time . (AREL) \\ \hline
Point-of-View Inconsistency & \textit{i was so excited to be graduating today . he was very proud of his graduation }. graduation day is always a success . he was very proud of his accomplishments . he was very proud of his accomplishments . (AREL) \\ \hline
\multirow{3}{*}{\begin{tabular}[c]{@{}l@{}}Excessive\\ Paraphrasing\end{tabular}} & we went on a trip to location . there were a lot of interesting things to see . there\textit{ were many different kinds of} fruits and vegetables . there\textit{ was also a variety of} fruits and vegetables . i had a great time there . (AREL) \\ \cline{2-2} 
 & we took the kids to  the park . \textit{we had a lot of fun . we had a great time .} the kids were having a great time . we had a great time . (Contextualize, Show and Tell)\\ \hline

\end{tabular}
\caption{Example stories associated with each error category.  We identify the system that predicted each example in parentheses, and indicate the specific component of the story in error in italics when applicable.}
\label{example_errors}
\end{table*}

We define our representative error categories as follows:
\begin{itemize}
    \item \textbf{Grammatical Errors:} Incorrect use of verbs and tenses and/or subject-verb disagreements.
    \item \textbf{Contradictions:} Presence of inconsistent ideas within the same story (e.g., two sub-stories that are the opposite of each other).
    \item \textbf{Repetitions}: These errors were further subdivided into the following categories.
    \begin{itemize}
        \item \textbf{Repetitions within Story:} Recurrence of the same sentence(s) within a story.
        \item \textbf{Repetitions within Sentence:} Recurrence of the same phrase(s) within a sub-story.
        \item \textbf{Repetitive Subject:} The sub-stories have the same subject and differ only in the adjective used to describe it. 
        \item \textbf{Repetitive Sentence Structure:} Most sentences start with ``the [noun] was/were/is [adjective].'' This leads to monotonous and unoriginal stories. We observed this error only in stories predicted by GLACNet.
    \end{itemize}
    \item \textbf{Description in Isolation:} Most sub-stories start with ``This is a picture of....'' Sentences of this nature sound more like single image captions than contextual stories.
    \item \textbf{Singular/Plural Disagreement:} The same story has one sentence with a singular noun and another sentence with the same noun but in plural form.
    \item \textbf{Ghost Entities:} Some sub-stories make use of a pronoun that has no antecedent at all (e.g., referring to a new person who was not introduced formally in the preceding sub-stories). This leads to confusion. 
    \item \textbf{Personification:} The attribution of human-like qualities to something non-human due to lack of common sense knowledge.
    \item \textbf{Absurdity:} Nonsensical stories or sub-stories. 
    \item \textbf{Incomplete Stories:} Stories that have less than 5 sentences. 
    \item \textbf{Point-of-View Inconsistency:} The narrative point of view randomly changes within the story (e.g., first person to second person), creating confusion.
    \item \textbf{Excessive Paraphrasing:} Presence of sub-stories that have similar meanings but are expressed using different words or phrases.
\end{itemize}

We provide examples of each of the above error types in Table \ref{example_errors}.  In addition to analyzing errors in stories with low predictions, we uncovered several anomalies in stories with high predictions.  We provide examples of these in Table \ref{scoring_anomalies}.

\begin{table}[t]
\centering
\small
\renewcommand{\arraystretch}{1.1}
\begin{tabular}{|m{0.75cm}|m{3.2cm}|m{2.35cm}|}
\hline
\textbf{Anom.} & \textbf{Example} & \textbf{Scores} \\ \hline
\multirow{2}{*}{\begin{tabular}[c]{@{}l@{}}Good\\ Story, \\ Low\\ Score\end{tabular}} & we went to a halloween party . there were a lot of interesting things to see . we saw a lot of cool things . we saw a lot of old buildings . the christmas tree was the best part of the day . (AREL) & CIDEr: 4.27, BLEU-4: 0.00, BLEU-3: 15.79, BLEU-2: 29.76, BLEU-1: 50.95, ROUGE-L: 24.43, METEOR: 24.42 \\ \cline{2-3} 
 & the couple was excited to be on vacation . they were going to the mountains . they went down the road . they saw a beautiful church . they had a nice dinner . (GLACNet) & CIDEr: 0.62 \\ \hline
\multirow{2}{*}{\begin{tabular}[c]{@{}l@{}}Bad\\ Story, \\ High\\ Score\end{tabular}} & the group of friends decided to go on a trip . they saw many interesting things . they stopped at a local restaurant . they had a great time . they ended up buying a new car . (GLACNet) & METEOR: 19.52, Bleu-4: 0.00, Bleu-3: 8.93, Bleu-2: 16.00, ROUGE-L: 22.55 \\ \cline{2-3} 
 & i went to a wedding last week . i had to take a picture of this beautiful flower . this is a picture of a woman . the flowers were so beautiful . the flowers were so beautiful . (AREL) & CIDEr: 20.90, Bleu-1: 71.79, Bleu-2: 43.47, METEOR: 33.98 \\ \hline
\end{tabular}
\caption{Example scoring anomalies, including the anomalous scores assigned to each story.}
\label{scoring_anomalies}
\end{table}

\begin{comment}
\subsubsection{Anomalies}
\begin{enumerate}
    \item Good story, Low score
\begin{itemize}
    \item 'we went to a halloween party . there were a lot of interesting things to see . we saw a lot of cool things . we saw a lot of old buildings . the christmas tree was the best part of the day .' (AREL)
    
    \textit{[CIDEr: 4.267338622, BLEU-4: 1.73E-03,	BLEU-3: 15.79219919, BLEU-2: 29.75630907, BLEU-1: 50.95039764, ROUGE-L: 24.4340596,	METEOR: 24.41528569]}
    \item 'the couple was excited to be on vacation . they were going to the mountains . they went down the road . they saw a beautiful church . they had a nice dinner .' (GLACNet)
    
    \textit{[CIDEr: 0.6152894139]}
\end{itemize}
    \item Bad story, high score
\begin{itemize}
    \item 'the group of friends decided to go on a trip . they saw many interesting things . they stopped at a local restaurant . they had a great time . they ended up buying a new car .' (GLACNet)
    
    \textit{[METEOR:19.52382228,
    Bleu-4:1.19E-03,
    Bleu-3: 8.926301665,
    Bleu-2: 16.00142241,
    ROUGE-L: 22.55083179]}
    \item ' i went to a wedding last week . i had to take a picture of this beautiful flower . this is a picture of a woman . the flowers were so beautiful . the flowers were so beautiful .' (AREL)
    
    \textit{[METEOR: 33.98146813,
    Bleu-1: 71.79487179,
    Bleu-2: 43.46652426,
    CIDEr: 20.90455481]}
\end{itemize}
\end{enumerate}
\end{comment}

\subsection{Discussion}
\label{error_discussion}

\begin{table}[t]
\small
\centering
\renewcommand{\arraystretch}{1.1}
\begin{tabularx}{\columnwidth}{|l|Y|Y|Y|}
\hline
\textbf{Error Category}                     & \textbf{AREL-s-50} & \textbf{\begin{tabular}[c]{@{}l@{}}GLAC-\\Net \end{tabular}} & \textbf{\begin{tabular}[c]{@{}l@{}}Contex-\\tualize, \\Show\\ and \\Tell\end{tabular}} \\ \hline
Repetition of Sub-Stories & 19.70\%           & 2.08\%      &15.42\%         \\ \hline
Description in Isolation   & 29.01\%          & 0\%  & 15.79\%               \\ \hline
\end{tabularx}
\caption{Frequency (in terms of overall percentage) of the most common error types across all 1010 generated test stories by AREL and GLACNet and 1938 generated test stories by Contextualize, Show and Tell.}
\label{error_statistics_table}
\end{table}

The most common error types we observed were repetitions and descriptions in isolation; we present statistics indicating the frequencies of these errors for AREL, GLACNet, and Contextualize, Show and Tell in Table \ref{error_statistics_table} (note that both occurred with the highest frequency in AREL).  The rarest error category was that containing incomplete stories.  This error appeared only in AREL stories, and only in three of the 1010 generated stories (0.003\%).

The prevalence of repetitions in AREL is likely a side-effect of the model's architecture---it generates the sub-stories for the whole album in parallel, rather than keeping track of what was generated in the previous sub-story. We found that this structure also led to some stories having contradictory sentences. In contrast, GLACNet stories exhibited few repetitions because of the post-processing step employed after decoding. In this step, words for a sentence are sampled from a word probability distribution one hundred times and the most frequent word is selected. The words which occur in the generated sentences are also counted and the selection probabilities of words are decreased as their frequency increases. 

It is somewhat surprising that the stories generated using Contextualize, Show and Tell also exhibited such a high frequency of repetitions, in spite of the fact that the model generated sub-stories sequentially. This demonstrates that some sort of feedback mechanism incorporating the model's previously generated sub-stories is needed. The output of each of the five decoders in Contextualize, Show and Tell should be fed into the next decoder to keep track of previously generated sub-stories.

We observed that there were very few grammatical errors in the GLACNet stories, as the probabilities associated with function words (e.g., prepositions and pronouns) remained unchanged even if their rate of occurrence was high. In contrast, stories generated by AREL (which includes no such grammar-checking mechanism) included a considerable number of grammatical errors.  GLACNet's post-processing step still could be improved upon---we were somewhat surprised to find that some of its stories used both singular and plural forms of the same noun within a story. We assume the error occurred due to the fact that the model decreases the probability of frequently occurring words. Thus, if the singular noun occurred in the previous sub-story, its plural form gets included in the next sub-story.

The within-sentence repetitions may at least partially be a consequence of the presence of repetitions in some VIST training stories. In our analysis of the crowdsourced dataset we found that human typing/grammar errors were a relatively common occurrence, resulting in imperfect training data. Although the stories generated by GLACNet did not often exhibit repetitions due to the reasons mentioned in the paragraph above, there was a trade-off in terms of originality of the generated stories.  We found that most were monotonous, using similar sentence structures for every story.

Descriptions in isolation, the single most prevalent error type we identified in AREL and Contextualize, Show and Tell stories, read more like image captions (describing the image's contents) than components of a sequential story. We are perplexed as to why these errors were so common, since to the best of our understanding the models did not include any DII instances in their training sets.  It may be the case that caption-like sub-stories are learned to be ``safer'' choices by these models, and thus generated more often than riskier contextual sub-stories.

Sentences that are lexically different but semantically similar cause redundancies in the story and are a common occurrence in both GLACNet and AREL.  Since images in a sequential album are often visually similar to one another, it may be the case that both models predict that two (or more) images in a sequence refer to the same content.  In attempting to vary the resulting sub-stories nonetheless, they succeed only at generating paraphrases of one another.

\section{Recommendations}
\label{recommendations}
As evidenced by our error analysis, there is substantial scope for improvement in visual storytelling. Based on our observations, we make the following recommendations.  First, \textbf{automatically preprocessing the DII and SIS training files remains an unexplored but potentially highly useful preliminary step in the story generation process.}  Doing so could aid future systems in avoiding grammatical mistakes, particularly if coupled with a post-processing mechanism similar to what is currently employed by GLACNet.  Second, in terms of the post-processing mechanism itself, \textbf{incorporating temporal sequencing methods will yield more well-organized and coherent stories}.  This could be done by sorting a (presumably jumbled) set of sub-stories after they have been generated, as was done by \citeauthor{agrawal-EtAl:2016:EMNLP2016} \shortcite{agrawal-EtAl:2016:EMNLP2016}.

Third, it is common for current models to generate all sub-stories in parallel. This leads to repetitions and redundancies in the generated stories. \textbf{Modifying the architecture in such a way that the sub-stories are generated sequentially and the word tokens of the previously generated sub-stories are passed back to the model may lead to numerous benefits.} For instance, this feedback could be used to identify past sub-story topics, as well as to ensure that the singularity/plurality of subjects remains the same across the entire story.  Incorporating a memory mechanism could also lessen the frequency of point-of-view inconsistencies, excessive paraphrasing, and contradictions. The architecture of the decoder used by \citeauthor{DBLP:journals/corr/VenugopalanRDMD15} \shortcite{DBLP:journals/corr/VenugopalanRDMD15} can also be adopted for providing feedback at the word level along with the sub-story level feedback. This will help in keeping track of the previously generated words in the story and prevent in-sentence repetitions.

Fourth, \textbf{traditional image captions (DIIs) can be (carefully) leveraged to support the generation of high-quality stories}, for instance by facilitating named entity recognition and thereby decreasing the frequency of ghost entities.  Another way to avoid ghost entities is to (fifth) \textbf{incorporate a bottom-up and top-down visual attention mechanism}, such as that used in prior image captioning work \cite{DBLP:journals/corr/AndersonHBTJGZ17}, to learn image-specific features and facilitate visual grounding.  Few-shot learning methods to jointly encode the images and text \cite{dong2018fpait} could also be used in this regard.

Sixth, although \citeauthor{DBLP:journals/corr/JainAMSLS17}'s \shortcite{DBLP:journals/corr/JainAMSLS17} work considered only textual features, \textbf{a machine translation model could be used} to produce more creative stories while avoiding repetitive sentence structures and absurdities to some degree. \citeauthor{DBLP:conf/eacl/MatusovWCSLC17} \shortcite{DBLP:conf/eacl/MatusovWCSLC17} use a neural machine translation model which contains a visual encoder and a textual encoder, thus giving attention independently to both image features and source sentences. This technique is a more viable option. Finally, the anomalies we uncovered in our error analysis validate the position first put forward by \citeauthor{DBLP:journals/corr/abs-1804-09160} \shortcite{DBLP:journals/corr/abs-1804-09160}, that automatic metrics leave much to be desired in terms of judging visual storytelling approaches.  We recommend that a standardized human evaluation metric be included in the assessment of these approaches in the future.

\section{Conclusion}
\label{conclusion}
In this work, we conduct a comprehensive error analysis of recent visual storytelling approaches.  We note current shortcomings in this area, and make recommendations for addressing these limitations in future work. We find that the most common errors are repetitions, the presence of traditional image descriptions, and a lack of creativity in the machine-generated stories. Preprocessing the training text, developing a combined visual and text co-attention mechanism, and sequentially generating sub-stories and providing them as feedback to the model could all help to ameliorate these issues.  Specifically, including these elements could help in the generation of more context-aware sequential sub-stories, and temporally sequencing the sub-stories will produce more creative, coherent, relevant, and most importantly, humanlike stories. We plan to experiment with the techniques mentioned above in our future work. 

\section*{Acknowledgements}
\label{ack}
We thank the anonymous reviewers for their comments and suggestions. This material is based upon work supported by Google Cloud.

\bibliography{naaclhlt2019}
\bibliographystyle{acl_natbib}

\end{document}